\documentclass[nohyperref]{article}

\usepackage{microtype}
\usepackage{graphicx}
\usepackage{subfigure}
\usepackage{booktabs} 

\usepackage{hyperref}

\usepackage[accepted]{icml2022}

\usepackage{mathtools}

\usepackage{amsmath,amsfonts,bm}

\def\eqref#1{equation~\ref{#1}}

\def\1{\bm{1}}

\DeclareMathAlphabet{\mathsfit}{\encodingdefault}{\sfdefault}{m}{sl}
\SetMathAlphabet{\mathsfit}{bold}{\encodingdefault}{\sfdefault}{bx}{n}

\usepackage{hyperref}
\usepackage{url}
\usepackage{graphicx}
\usepackage{amsmath}
\usepackage{bbding}
\usepackage{pifont}
\usepackage{wasysym}
\usepackage{amssymb}
\usepackage{amsthm}
\usepackage{bm}
\usepackage{booktabs}
\usepackage{multirow}
\usepackage{caption}
\usepackage{tabularx}

\usepackage{comment}

\usepackage{float}
\usepackage{wrapfig,lipsum,booktabs}
\newfloat{figtab}{htb}{fgtb}
\makeatletter
  \newcommand\figcaption{\def\@captype{figure}\caption}
  \newcommand\tabcaption{\def\@captype{table}\caption}
\makeatother

\usepackage[capitalize,noabbrev]{cleveref}
\theoremstyle{plain}
\theoremstyle{definition}

\theoremstyle{remark}

\usepackage[textsize=tiny]{todonotes}

\icmltitlerunning{Your  Autoregressive  Generative  Model  Can  be Better If You Treat It as an Energy-Based One}

\begin{document}

\twocolumn[
\icmltitle{Your  Autoregressive  Generative  Model  Can  be Better If You Treat It as an Energy-Based One}



\icmlsetsymbol{equal}{*}

\begin{icmlauthorlist}
\icmlauthor{Yezhen Wang}{equal,yyy}
\icmlauthor{Tong Che}{equal,yyy}
\icmlauthor{Bo Li}{com}
\icmlauthor{Kaitao Song}{sch}
\icmlauthor{Hengzhi Pei}{zzz}
\icmlauthor{Yoshua Bengio}{yyy,cifar}
\icmlauthor{Dongsheng Li}{sch}
\end{icmlauthorlist}

\icmlaffiliation{yyy}{Mila-Quebec AI Institute / U. Montreal}
\icmlaffiliation{cifar}{CIFAR Senior Fellow and AI Chair}
\icmlaffiliation{com}{S-Lab, Nanyang Technological University}
\icmlaffiliation{sch}{Microsoft Research Asia, China}
\icmlaffiliation{zzz}{University of Illinois, Urbana Champaign}
\icmlcorrespondingauthor{Yezhen Wang}{yezhen.wang0305@gmail.com}
\icmlcorrespondingauthor{Tong Che}{tongcheprivate@gmail.com}


\vskip 0.3in
]



\printAffiliationsAndNotice{\icmlEqualContribution} 

\begin{abstract}
Autoregressive generative models are commonly used, especially for those tasks involving sequential data. They have, however, been plagued by a slew of inherent flaws due to the intrinsic characteristics of chain-style conditional modeling (e.g., exposure bias or lack of long-range coherence), severely limiting their ability to model distributions properly. In this paper, we propose a unique method termed E-ARM for training autoregressive generative models that takes advantage of a well-designed energy-based learning objective. By leveraging the extra degree of freedom of the softmax operation, we are allowed to make the autoregressive model itself be an energy-based model for measuring the likelihood of input without introducing any extra parameters. Furthermore, we show that E-ARM can be trained efficiently and is capable of alleviating the exposure bias problem and increase temporal coherence for autoregressive generative models. Extensive empirical results, covering benchmarks like language modeling, neural machine translation, and image generation, demonstrate the effectiveness of the proposed approach.

\end{abstract}
\section{Introduction}

By factorizing the joint distribution into the product of a series of conditional distributions,  autoregressive generative models (abbr. ARGMs)~\citep{TransformerBase, TransformerXL, wavenet, pixelcnnOordKEKVG16, pixelcnn++SalimansK0K17, PixelSNAILChenMRA18} simplify the difficult challenge of modeling high-dimensional joint distributions. They can be trained efficiently via maximum likelihood and generate samples of exceptional quality, making this technique popular for modeling distributions, especially for sequential data. Nonetheless, despite their potency and flexibility, ARGMs still have inherent weaknesses due to the intrinsic characteristics of chain-style conditional modeling. For example, ARGMs usually suffer from a discrepancy of the input context distributions between the training and inference stages, which causes consequent error propagation (\textit{i.e.}, Exposure Bias~\citep{SSRanzatoCAZ15, SSBengioVJS15}). Besides, due to the nature of greedy selection of beam search approximations, the decoded results from ARGMs may also lack in long-range coherence.  We consider one approach by which ARGMs could be adapted to reduce these concerns. 

Earlier work, both heuristic and theoretical, has already been proposed with those goals. For instance, the exposure bias problem of ARGMs can be alleviated to some extent with scheduled sampling~\citep{SSBengioVJS15, SSTransformerMihaylovaM19}, by mixing input contexts from both real data and autoregressive generation, during the training stage. However, this scheme suffers from an over-correcting problem~\citep{BridginfGapNMTijcai}.  In addition, at the inference stage, beam search makes it possible to choose more diverse candidates, improving the quality of generated sequences. Nevertheless, this results in only marginal improvements in temporal coherence, since ARGMs can only leverage previous decoded contexts without consideration of the whole sequence information. Moreover, setting aside the difficulty in training them, energy-based models (EBMs) have demonstrated their effectiveness in modeling high-dimensional distributions in a variety of machine learning applications~\citep{zhao2017energy,arbel2021generalized, gao2021learning}, without requiring the transformation of the target distribution into a product of conditional distributions. As a result, several studies~\citep{residualDengBOSR20, ResidualBakhtinDGORS21, autoregressiveEnergyDurkanN19} attempt to combine EBMs with ARGMs, expecting to benefit from the strengths of both approaches. However, though some positive results were obtained, the existing works preferred a two-stage optimization, which first obtains a well-trained ARGM and then trains an additional EBM based on it. Such an optimization strategy does not enable ARGMs itself to benefit from the properties of EBM in modeling the joint distribution in a temporally more coherent way, and requires more training parameters to estimate energy scores, burdening the intricacy of the learning task.

In this paper, we present a novel design, which seamlessly integrates {\tt\bf E}nergy{\tt\bf -}based models into {\tt\bf A}uto{\tt\bf R}egressive {\tt\bf M}odels (E-ARM) by utilizing the extra degree of freedom within the model's final softmax layer. We will show that in this way the ARGM can be trained using an energy-based learning objective, which allows the ARGM to not only avoid those intrinsic concerns, such as exposure bias, with the help of energy-based models as former work did~\citep{residualDengBOSR20, ResidualBakhtinDGORS21}, but also be free of increasing the learning model's complexity. This property makes our E-ARM rather easy to be applied on the training process of any autoregressive generative model for any specific task, as no structural changes are required. 

Besides, we follows the predominant approach for training explicit density(mass) generative models to minimize the KL divergence between the (empirical) data distribution and model distribution, which gives rise to the gradient-based contrastive divergence methods~\citep{Hinton02Training, DeepDirectedKimB16} for energy-based models. Typically, these methods require an Markov Chain Monte Carlo (MCMC) process to sample data from the EBM for the ``negative phase'' gradient estimation, which is extremely time-consuming and, meanwhile, inapplicable for discrete data, such as text. To solve this, we present a way to estimate those ``negative phase'' gradients through those samples generated with the network's autoregressive view instead of the EBM view, which allows us to sidestep the usage of MCMCs, thus making the training both feasible and efficient.

Intuitively, the exposure bias in ARGMs is caused by the fact that the model is trained on real data rather than data generated by the model. On the other hand, in the EBM’s optimization process for modeling joint densities, the negative phase of contrastive divergence methods~\citep{Hinton02Training, DeepDirectedKimB16} requires sampling data from the EBM itself. Along with the fact that our method combines the EBM and the ARGM seamlessly as a whole, E-ARM can reduce the discrepancy between input data of the training and inference stage, which mitigates the exposure bias problem of the ARGM. On top of it, unlike ARGMs, which factor the joint distribution into a product of conditional distributions, EBMs are able to model the joint distribution directly and score each input at the sequence level instead of at the token level, which makes them capable of modeling long-range coherence. 

In summary, the following contributions are made with this paper: i) We introduce a novel scheme, E-ARM, to integrate the EBM view into autoregressive generative models seamlessly; ii) we attempt to reduce the intrinsic problems of autoregressive models, such as exposure bias and weak temporal coherence, by optimizing an energy-based learning objective, which uses samples autoregressively generated; iii) We demonstrate how to efficiently optimize our model constructed from a single network, using the contrastive divergence method without MCMC; iv) In a number of applications, such as language modeling, neural machine translation, our model can achieve better results in comparison with relevant baselines.

\section{Background}
\label{sec:Background}
\subsection{Energy-Based Models}
Let $P_d$ and $P_\theta$ denote the data distribution and the model distribution with density(mass) functions $p_d$ and $p_\theta$ with respect to a base measure $\bf{d} x$ on sample space $\mathcal{X} \subset \mathbb{R}^{m}$ respectively. Energy-based models~\citep{lecun2006tutorial} are interested in learning an unnormalized energy function $\mathbf{E}_{\theta}(\mathbf{x})$ that defines the density(mass) function $p_\theta(\mathbf{x})$ of the model distribution $P_\theta$,
\begin{equation} \label{eq:ebm_density}
p_\theta(\mathbf{x}) = \frac{\exp(- \mathbf{E}_{\theta}(\mathbf{x}))}{\mathbf{Z}_\theta},
\end{equation}
where $E_{\theta}: \mathcal{X} \to \mathbb{R}$ denotes an energy function which aims to map a data sample to an energy scalar, and $\mathbf{Z}(\theta) = \sum_\mathbf{x} \exp(- \mathbf{E}_{\theta}(\mathbf{x}))$ denotes the normalizing constant, also known as the partition function. Any function can be used as an energy function to represent an EBM as long as it can generate a single scalar given some input $\mathbf{x}$ and the normalizing constant is finite\footnote{Without constraining the parametrization of $\mathbf{E}_{\theta}$, this can be achieved by bounding the region of space in which $x$ takes its allowed values.}. Contrastive divergence algorithms are commonly used to optimize EBMs~\citep{Hinton02Training, DeepDirectedKimB16,jemGrathwohlWJD0S20} via maximum log-likelihood. Correspondingly, the gradient of the log-likelihood, which needs to be maximized, with respect to $\theta$ can be expressed as 
\begin{gather}
\mathbb{E}_{P_d(\mathbf{x})}\Big[\nabla_{\theta} \log p_\theta(\mathbf{x})\Big] \nonumber  \\ = \mathbb{E}_{P_\theta(\mathbf{x})}\Big[ \nabla_{\theta} \mathbf{E}_\theta (\mathbf{x}) \Big] - \mathbb{E}_{P_d(\mathbf{x})}\Big[ \nabla_{\theta} \mathbf{E}_\theta (\mathbf{x}) \Big].
\label{eq:derivative_log_likelihood}
\end{gather}
The first term in the right hand side of Eq.\ref{eq:derivative_log_likelihood} is usually called ``negative phase'' term while the second term is called ``positive phase'' term. In general, it is non-trivial to sample from an EBM, which usually requires MCMC methods~\citep{Hinton02Training, SGLDWellingT11}. Stochastic Gradient Langevin Dynamics~\citep{SGLDWellingT11} is the most common MCMC algorithm in continuous state spaces. However, they are exceedingly time-consuming and not applicable when the input is discrete like in text applications.

\subsection{Modeling Distributions Autoregressively}
Autoregressive generative models can decompose any joint distribution into a product of conditional distributions using the product rule of probability by ordering those random variables within the joint distribution and characterizing each random variable given all variables preceding it in that order. Formally, we use $\mathbf{x}_{< k}$ to denote the random vector covering all random variables before the time step $k$ and $\mathbf{x}_k$ denote the random variable at time step $k$. Then we have
\begin{equation}
\begin{aligned}
p(\mathbf{x}) = \prod^{K}_{k=1} p(\mathbf{x}_k | \mathbf{x}_{< k}).
\end{aligned}    
\end{equation}
In recent years, remarkable accomplishments in numerous areas has been achieved by modeling distributions autoregressively~\citep{TransformerBase, GPT2, pixelrnn, pixelcnnOordKEKVG16, pixelcnn++SalimansK0K17}, thanks to its ability to avoid the challenging goal of modeling joint high-dimensional distributions directly. In this paper, we primarily focus on autoregressive language models, but we also conduct experiments on image generation to validate the generality of our method.

\subsection{Exposure Bias and Incoherence Problems in Autoregressive Models}
In the discussion about the defects of sequential autoregressive generative models, the exposure bias problem~\citep{SSBengioVJS15, SSRanzatoCAZ15} is an important issue, which greatly affects the model's deployment performance. During the training stage, the autoregressive model is always conditioned on ground truth token sequences. During the inference stage, however, the model has to rely on its own previously generated tokens to predict the next token, when the model is deployed. If an incorrect token is selected, this error can be amplified in following steps because the next prediction will be made using an unusual input (one unlike those in the training set). Besides, out of the consideration of efficiency, autoregressive decoding usually selects the most probable token at each time step, given the ones previously selected. Such a scheme assumes the largest joint probability of the whole sequence can be achieved by separately choosing the most probable next token (given its previous context) over all time steps, which is only the local optimum. Correspondingly, the chosen sequence can not always be the model's optimum result.

\section{Methodology}

For a long time, as a result of compromises for improving training stability and efficiency (\textit{e.g.}, modeling a joint distribution by decomposing it and using a teacher-forcing training strategy), conventional autoregressive generative models have suffered from flaws such as the exposure bias and the lack of long-range coherence. To tackle these issues, we attempt to seamlessly integrate {\tt\bf E}nergy{\tt\bf -}based models into {\tt\bf A}uto{\tt\bf R}egressive {\tt\bf M}odels (E-ARM), and train ARGMs with an energy-based learning objective. 

\label{sec:model}
Formally, given a distribution $P_d(\mathbf{x}_k, \mathbf{x}_{<k})$ of length $k$ sequences with density(mass) function $p_d(\mathbf{x}_k, \mathbf{x}_{<k})$ on data space $\mathcal{X}^{k}$, we first introduce a parametric autoregressive model $Q_\theta$ with a density(mass) function $q_\theta(\mathbf{x}_k, \mathbf{x}_{<k})=\prod_{l=1}^{k}q_\theta(\mathbf{x}_l|\mathbf{x}_{< l})$ with parameter $\theta$. Then, we define the energy-based autoregressive model $P_\theta(\mathbf{x}_k, \mathbf{x}_{<k})$, which is a product of the autoregressive model and an EBM adhered within it, with density function $p_\theta(\mathbf{x}_k, \mathbf{x}_{<k})$,
\begin{equation}
\begin{aligned}
\label{eq:prod_ebm}
p_\theta(\mathbf{x}_k, \mathbf{x}_{<k}) = q_\theta(\mathbf{x}_{< k}) \cdot \frac{e^{-\phi_{\theta}(\mathbf{x}_k, \mathbf{x}_{< k})}}{\mathbf{Z}_\theta},
\end{aligned}
\end{equation}
where $\mathbf{Z}_\theta$ is the normalization term  and equals to $\mathbb{E}_{\mathbf{x}_{< k} \sim Q_\theta(\mathbf{x}_{< k})}[ \sum_{\mathbf{x}_k}  e^{-\phi_{\theta}(\mathbf{x}_k, \mathbf{x}_{< k})}]$, the energy function $\phi_\theta(\mathbf{x}_k, \mathbf{x}_{< k})$ is defined as the negative of $\mathbf{x}_K$'s corresponding component of network's output logits given the input context $\mathbf{x}_{< k} = (\mathbf{x}_1, \mathbf{x}_2, \dots, \mathbf{x}_{k-1})$ (\textit{e.g.}, given a sequence ``This is Friday.'' and assuming the corresponding index of the token ``Friday'' in the vocabulary is $i$, then the value of $- \phi_\theta(\text{``Friday'', ``This is''})$ is the $i$-th component of the output logit, namely, the input tensor of the final softmax layer). 

One rationale behind such a design is out of the extra degree of freedom concealed inside the softmax operation. Specifically, a $M$-way softmax operation is a transformation $\mathcal{S}: \mathbb{R}^M \rightarrow \mathbb{R}^M$, which can convert an unnormalized vector into a normalized vector with 1 as the sum of all elements.
\begin{equation}
\begin{aligned}
\mathcal{S}([z_1, \dots, z_M]) = [\frac{e^{z_1}}{\sum_{i=1}^{M}e^{z_i}}, \dots, \frac{e^{z_M}}{\sum_{i=1}^{M}e^{z_i}}],
\end{aligned}
\end{equation}
where $z_i \in \mathbb{R}$. It's easy to observe that the softmax operation is unaffected by the input vector's overall magnitude, that is, $\mathcal{S}([z_1, \dots, z_M]) = \mathcal{S}([z_1, \dots, z_M] + C), C \in \mathbb{R}$, indicating an extra degree of freedom which can be used for modeling energy.

Our primary goal is to make the parametric distribution $P_\theta(\mathbf{x}_k, \mathbf{x}_{< k})$ approach the real data distribution $P_d(\mathbf{x}_k, \mathbf{x}_{< k})$ as close as possible at any time step $k \leq K$ such that we can perform downstream inferences (e.g., density evaluation and sampling) via $P_\theta$. This can be achieved by minimizing the Kullback-Leibler (KL) divergence between the distributions with respect to all time steps of a sequence,
\begin{gather}
\label{eq:KL_divergence}
\min\limits_{\theta} \sum\limits_{k=1}^{K} \bigg[ \lambda_k \textbf{D}_{KL}\Big(P_d(\mathbf{x}_k, \mathbf{x}_{< k}) || P_\theta(\mathbf{x}_k, \mathbf{x}_{< k})\Big) \bigg],
\end{gather}
where $\lambda_k$ adjusts the weights of objectives with respect to different time steps, though we found that just setting all $\lambda_k$ equal to each other could work well. We resort to contrastive divergence methods~\citep{Hinton1995TheA, DeepDirectedKimB16} to minimize the objective~\ref{eq:KL_divergence} by descending the gradient w.r.t. $\theta$ according to Eq.~\ref{eq:derivative_log_likelihood}\footnote{here, we take a minimization version of the  Eq.~\ref{eq:derivative_log_likelihood}. Thus the sign before each phase  is converse.} for all time steps. For a specific time step $k$, we have the gradient $\mathcal{G}^{k}(\theta)$
\begin{gather}
\mathcal{G}^{k}(\theta) = \mathcal{G}^{k}_{+}(\theta) - \mathcal{G}^{k}_{-}(\theta) \label{eq:pos_neg_gradient}, \\
\mathcal{G}^{k}_{+}(\theta) = \mathbb{E}_{\mathbf{x}_k, \mathbf{x}_{< k} \sim P_d(\mathbf{x}_k, \mathbf{x}_{< k})} \bigg[\nabla_{\theta} \mathbf{E}_\theta(\mathbf{x}_k, \mathbf{x}_{< k})\bigg] \label{eq:pos_gradient},  \\
 \mathcal{G}^{k}_{-}(\theta) =\mathbb{E}_{\mathbf{x}_k, \mathbf{x}_{< k} \sim P_\theta(\mathbf{x}_k, \mathbf{x}_{< k})}\bigg[\nabla_{\theta} \mathbf{E}_\theta(\mathbf{x}_k, \mathbf{x}_{< k}) \bigg] \label{eq:neg_gradient},
\end{gather}
where $\mathbf{E}_\theta(\mathbf{x}_k, \mathbf{x}_{< k}) = \phi_{\theta}(\mathbf{x}_k, \mathbf{x}_{< k}) - \log q_\theta(\mathbf{x}_{< k})$. Optimization via Eq.~\ref{eq:pos_neg_gradient} involves sampling data from the model distribution $P_\theta$ and can thus lead to the discovery of non-data-like samples, whose likelihood is then explicitly reduced as the corresponding energy increasing during the training. E-ARM is therefore not plagued by the exposure bias problem naturally. Besides, because we model the joint distribution at each time step throughout the training process, E-ARM can assess the entire sequence as a whole and generate more coherent data using energy sampling~\citep{residualDengBOSR20}.
\section{Optimization}
\label{sec:optimization}
The key obstacle of optimizing the objective ~\ref{eq:KL_divergence} via contrastive divergence methods is sampling data from the model distribution $P_\theta$ for estimating the ``negative phase'' gradient $\mathcal{G}^{k}_{-}(\theta)$ since we can only access the estimated density(mass) function $p_\theta$. The common MCMC algorithms are not desirable for generating ``negative'' samples because they are rather time-consuming, and not applicable for discrete data such as text. In order to make the optimization process both efficient and feasible, we propose a unique way to conduct the optimization by means of importance sampling technique~\cite{importancesampling}.
To be specific, by replacing $\mathbf{E}_\theta(\mathbf{x}_k, \mathbf{x}_{< k})$ with the specific form $\phi_{\theta}(\mathbf{x}_k, \mathbf{x}_{< k}) - \log q_\theta(\mathbf{x}_{< k})$ in Eq.\ref{eq:pos_gradient}, the ``positive phase'' gradient $\mathcal{G}^{k}_{+}(\theta)$ with respect to parameter $\theta$ can be written into
\begin{equation}
\begin{aligned}
\label{eq:wake_phase}
 \mathcal{G}^{k}_{+}(\theta) =  \mathbb{E}_{P_d}\Big[\nabla_{\theta} \phi_{\theta}(\mathbf{x}_k, \mathbf{x}_{< k}) - \nabla_{\theta} \log q_\theta (\mathbf{x}_{< k})\Big],
\end{aligned}    
\end{equation}
and similarly, we can get the ``negative phase'' gradient 
\begin{equation}
\begin{aligned}
\label{eq:sleep_phase}
 \mathcal{G}^{k}_{-}(\theta) =  \mathbb{E}_{P_\theta}\Big[\nabla_{\theta} \phi_{\theta}(\mathbf{x}_k, \mathbf{x}_{< k}) - \nabla_{\theta} \log q_\theta (\mathbf{x}_{< k})\Big].
\end{aligned}    
\end{equation}
We can observe that the term $\mathbb{E}_{P_d}[- \nabla_{\theta}\log q_\theta (\mathbf{x}_{< k})]$ of $\mathcal{G}^{k}_+(\theta)$ in  Eq.~\ref{eq:wake_phase} is equivalent to the negative gradient of likelihood $q_\theta(\mathbf{x}_{< k})$'s logarithm, which is exactly the objective of maximizing the autoregressive generative model $Q_\theta$'s log-likelihood, and can be easily taken care of via cross entropy loss.
Besides, because carrying out sample estimation of the expectation over the data distribution $P_d$ is viable, and the score $\phi_\theta(\mathbf{x}_k, \mathbf{x}_{< k})$ can be acquired by simply accessing the output logit of ARGM (according to the definition of $\phi_\theta$ in Sec. \ref{sec:model}), the ``positive phase'' gradient $\mathcal{G}^{k}_{+}$ can likewise be readily estimated.

The negative phase gradient estimation, on the other hand, is more involved. In Eq.~\ref{eq:sleep_phase}, sampling data from $P_\theta$ is required for estimating the expectation $\mathbb{E}_{P_\theta}$, whereas $P_\theta$ is the introduced energy-based autoregressive model, which is an explicit autoregressive generative model and we can only access its modeled density(mass) function $p_\theta$. However, inspired by the idea of importance sampling, we substitute the troublesome estimation of the expectation over distribution $P_\theta$ with the expectation over distribution $Q_\theta$, which is the underlying autoregressive model that can generate samples considerably easier, in practice. Accordingly, the ``negative phase'' gradient $\mathbb{E}_{\mathbf{x}_k, \mathbf{x}_{< k}\sim P_\theta}[\nabla_{\theta} \mathbf{E}_\theta(\mathbf{x}_k, \mathbf{x}_{< k})] $ has the following form (See the detailed derivation in Appendix \ref{appendix:A}),
\begin{gather}
\mathcal{G}^{k}_{-}(\theta) =  \mathbb{E}_{Q_\theta(\mathbf{x}_k, \mathbf{x}_{< k})}\Big[\mathbf{w}(\mathbf{x}_{<k})\nabla_{\theta}\phi_\theta(\mathbf{x}_k, \mathbf{x}_{< k})\Big] \nonumber \\ \ \  -\mathbb{E}_{Q_\theta (\mathbf{x}_{< k}) }\Big[\mathbf{w}(\mathbf{x}_{< k})\nabla_{\theta} \log q_\theta (\mathbf{x}_{< k})\Big] \label{eq:sleep_phase_derived},
\end{gather}
where
\begin{gather}
\mathbf{w}(\mathbf{x}_{< k})  =  \frac{\sum_{\mathbf{x}_k}e^{-\phi(\mathbf{x}_k, \mathbf{x}_{< k})}}{\mathbb{E}_{\mathbf{x}'_{< k} \sim Q_\theta(\mathbf{x}_{< k})}[ \sum_{\mathbf{x}_k}  e^{-\phi_{\theta}(\mathbf{x}_k, \mathbf{x}'_{< k})}]}\label{eq:w(x)}.
\end{gather}
According to Eq.\ref{eq:sleep_phase_derived}, all the estimated expectations only need sampling from the autoregressive model $Q_\theta$ rather than the distribution $P_\theta$, and the reweighing weight $\textbf{w}$ in  Eq.~\ref{eq:w(x)} does not involve expectation computation over distribution $P_\theta$ either. Generally speaking, producing data from an autoregressive model is a simple ancerstral sampling process and naturally suitable for discrete data, as compared with sampling straight from an explicit generative density(mass) estimator, which needs MCMC approaches~\citep{autoregressiveEnergyDurkanN19}. Besides, the term $\mathbb{E}_{\mathbf{x}_{< k} \sim Q_\theta (\mathbf{x}_{< k}) }[\mathbf{w}(\mathbf{x}_{< k})\nabla_{\theta} \log q_\theta (\mathbf{x}_{< k})  ] $ in Eq. \ref{eq:sleep_phase_derived} can be regarded as a re-weighted gradient of $Q_\theta$'s information entropy with respect to $\theta$. This term can be optimized similarly to the teacher-forcing training of autoregressive model with the ``teacher" sequence generated autoregressively by the model itself. Actually, the scheduled sampling methods~\citep{SSBengioVJS15, SSRanzatoCAZ15, SSTransformerMihaylovaM19} are similar to this term but without the re-weighting factor. 
Moreover, the reweighing weight $\mathbf{w}$ of Eq.~\ref{eq:w(x)} can be further refined (see the derivation in Appendix \ref{sec:appendix_w}) and we can observe that
\begin{equation}
\begin{aligned}
\mathbf{w}(\mathbf{x}_{< k})  =\frac{\mu(\mathbf{x}_{< k})}{\mathbb{E}_{\mathbf{x}'_{< k}} \mu(\mathbf{x}_{< k})},
\end{aligned}    
\end{equation}
where $\mu (\mathbf{x}_{< k}) =\frac{p_\theta(\mathbf{x}_{< k})}{q_\theta(\mathbf{x}_{< k})}$, indicating the possibility of which distribution ($P_\theta$ or $Q_\theta$) the input context $\mathbf{x}_{< k}$ is most likely to come from. Correspondingly, $\textbf{w}(\mathbf{x}_{< k})$ reflects the context $\mathbf{x}_{< k}$'s relative magnitude of $\mu (\mathbf{x}_{< k})$ compared with the average among all potential contexts---the larger the value of $\textbf{w}(\mathbf{x}_{< k})$, the more likely the context $\mathbf{x}_{< k}$ in the data space coming from $P_\theta$, which is modeled by the product of autoregressive models and EBMs. During training, those input sequences with contexts more likely from $P_\theta$ than $Q_\theta$ will be assigned larger weights $\textbf{w}$ while others will be assigned smaller weights $\textbf{w}$. A precise sample estimate of the denominator of $\textbf{w}$ is important; otherwise, the gradient estimate $\mathcal{G}^{k}_{-}(\theta)$ will be biased. Fortunately, the bias can be reduced by replacing the estimate by an exponential moving average as MINE~\citep{mine} did. For small learning rates, this improved gradient estimation can be made to have arbitrarily small bias.

Ultimately, combining  Eq.~\ref{eq:wake_phase} and  Eq.~\ref{eq:sleep_phase_derived} , at each time step $k$, we can optimize $P_\theta(\mathbf{x}_k, \mathbf{x}_{< k})$ via descending the estimated gradient of $\theta$ as follows,
\begin{gather}
\mathcal{G}^{k}(\theta) = \Bigg(\underbrace{
\begin{aligned}
&- \ \ \mathbb{E}_{\mathbf{x}_{< k} \sim P_d(\mathbf{x}_{< k})}\Big[\nabla_{\theta}\log q_\theta (\mathbf{x}_{< k}) \Big] \\&+\mathbb{E}_{\mathbf{x}_k, \mathbf{x}_{< k}\sim P_d(\mathbf{x}_k, \mathbf{x}_{< k})}\Big[\nabla_{\theta}\phi_{\theta}(\mathbf{x}_k, \mathbf{x}_{< k})\Big]
\end{aligned}
}_{\textbf{Positive Phase}} \Bigg) \nonumber \\
-
\Bigg(\underbrace{
\begin{aligned}
&- \ \ \ \ \mathbb{E}_{\mathbf{x}_{< k} \sim Q_\theta (\mathbf{x}_{< k}) }\Big[\mathbf{w}(\mathbf{x}_{< k})\nabla_{\theta} \log q_\theta (\mathbf{x}_{< k})  \Big]\\
&+ \mathbb{E}_{\mathbf{x}_k, \mathbf{x}_{< k}\sim Q_\theta(\mathbf{x}_k, \mathbf{x}_{< k})}\Big[\mathbf{w}(\mathbf{x}_{< k})\nabla_{\theta}\phi_\theta(\mathbf{x}_k, \mathbf{x}_{< k})\Big]
\end{aligned}
}_{\textbf{Negative Phase}}\Bigg)\label{eq:final_objective}.
\end{gather}
Eq.~\ref{eq:final_objective} is a rather symmetric form that can be easily estimated by using ``positive'' samples from the given dataset distribution $P_d$ and ``negative'' samples from the base autoregressive model $Q_\theta$. 

In general, E-ARM can be viewed as a new training method for autoregressive models, which regards the ARGM as an EBM and can be optimized by a modified contrastive divergence method. 
The modeling of the joint distribution $P_d(\mathbf{x}_k, \mathbf{x}_{< k})$ instead of the conditional one at each time step $k$ ensures the autoregressive network stays close to the real distribution $P_d$ while avoiding those problems such as exposure bias and the lack of long-range coherence. However, in practice, training the model from scratch with the energy-based learning objective using gradients descending through Eq.\ref{eq:final_objective} alone can not work well. 
The reason is that at the initial stage of the training process, what we have is just a randomly initialized autoregressive network which outputs sequences with random values given any context. This indicates disjoint supports between the real sequence's distribution $P_d$ and distribution $P_\theta$ modeled by E-ARM. If we only use the energy-based learning objective derived according to contrasitve divergence methods, the gradient of log-likelihood $\mathbb{E}_{P_d(\mathbf{x})}[\nabla_{\theta} \log p_\theta(\mathbf{x})]$ would be nearly zero due to disjoint supports, which makes the optimization difficult. As a result, in order to make the optimization more feasible, we train the base autoregressive model with cross entropy loss for a few epochs before introducing the energy-based learning objective with respect to $P_\theta$. Actually, we set the starting epoch of E-ARM objective as a hyper-parameter, see more experimental details in the Section \ref{sec:app_lm}.

On top of it, it is worth noting that for a sequence with total length $K$, the gradient $\mathcal{G}^{k}(\theta)$ actually has $K$ counterparts with different time steps and the joint distribution $Q_\theta (\mathbf{x}_{< k})$ modeled by an autoregressive model can be naturally broken up into pieces of conditionals. As a consequence, simply summing up these $K$ gradients results in the ``negative phase'' gradient $\mathcal{G}_{-} = \sum_{i=1}^{K}\mathcal{G}^k_{-}$ having one term as follows
\begin{gather}
\sum_{k=1}^{K}\mathbb{E}_{ Q_\theta (\mathbf{x}_{< k}) }[\mathbf{w}(\mathbf{x}_{< k})\nabla_{\theta} \log q_\theta (\mathbf{x}_{< k})  ] \nonumber \\ =  \sum_{l=1}^{K} \sum_{k=1}^{K+1-l}\mathbb{E}_{ Q_\theta (\mathbf{x}_{< k})}[\mathbf{w}(\mathbf{x}_{< k})\nabla_{\theta} \log q_\theta ( \mathbf{x}_l| \mathbf{x}_{< l})],
\end{gather}
where $l$ indicates the specific index of the current token in the entire sequence. As a result, earlier time steps (smaller $l$) will get stronger training signals (larger $K+1-l$, indicating more gradient terms), giving rise to imbalanced training for different time steps. To solve this, we substitute  $\tilde{q}_\theta(\mathbf{x}_{< k})=\prod_{l=m}^{k-1}q_\theta(\mathbf{x}_l|\mathbf{x}_{< l})\prod_{n=1}^{m-1}q(\mathbf{x}_n|\mathbf{x}_{< n})$ for $q_\theta(\mathbf{x}_{<k})=\prod_{l=1}^{k}q_\theta(\mathbf{x}_l|\mathbf{x}_{< l})$ to define $p_\theta(\mathbf{x}_k, \mathbf{x}_{< k}) \propto \tilde{q}_\theta(\mathbf{x}_{< k})\cdot \exp (-\phi_{\theta}(\mathbf{x}_k, \mathbf{x}_{< k}))$, where $q(\mathbf{x}_n|\mathbf{x}_{< n})$ is a fixed copy of $q_\theta(\mathbf{x}_n|\mathbf{x}_{< n})$. Such a design constrains gradients only backpropagate through conditional distributions of a few recent tokens so that balances the training signals among different time steps \footnote{In practice, we find that using recent 2 tokens worked best.}.

\section{Experiments}
To empirically corroborate the effectiveness of E-ARM and show its broad applicability, we have conducted extensive experiments and in this section, we will introduce these experimental setups, followed by an analysis of the obtained results. We mainly focus on applications of natural language processing, but also carry out some simple experiments on image generation task to show that our E-ARM is a general training method for autoregressive models. More experimental settings and analytical experiments are shown in Appendix \ref{sec:expr_setting} and \ref{sec:more_experiment}.

\subsection{Application to Neural Machine Translation}
\begin{table*}[htp]
\setlength{\abovecaptionskip}{0.2cm}
\setlength{\belowcaptionskip}{0.0cm}
\centering
\renewcommand{\arraystretch}{1.2}
\resizebox{\textwidth}{!}{%
\begin{tabular}{c|cc|c|cccccc|c}
\toprule
\multirow{2}{*}{\textbf{Model}}  &
\textbf{Label} & \textbf{Scheduled} & \textbf{Beam} &
\multicolumn{6}{c|}{\textbf{BLEU} \bm$\uparrow$}  & \multirow{2}{*}{\textbf{Avg.}} \\
\cline{5-10}
& \textbf{Smoothing} & \textbf{Sampling} & \textbf{Searching} & \textbf{DE$\rightarrow$ EN} & \textbf{EN$\rightarrow$ DE} & \textbf{EN$\rightarrow$ IT} & \textbf{IT$\rightarrow$ EN} & \textbf{ES$\rightarrow$ EN} & \textbf{EN$\rightarrow$ ES} & \\
\midrule[1.2pt]
\multirow{6}{*}{\textbf{Base}} & \multirow{2}{*}{-} & \multirow{2}{*}{-} & - & 32.44\footnotesize$\pm$\footnotesize 0.06 & 26.64\footnotesize$\pm$\footnotesize 0.10 & 27.92\footnotesize$\pm$\footnotesize 0.03 & 30.48\footnotesize$\pm$\footnotesize 0.08 & 38.61\footnotesize$\pm$\footnotesize 0.11 & 35.42\footnotesize$\pm$\footnotesize 0.09 & 31.92 \\
&  &  & \textbf{5 B} & 33.62\footnotesize$\pm$\footnotesize 0.07  & 27.41\footnotesize$\pm$\footnotesize 0.08  & 28.72\footnotesize$\pm$\footnotesize 0.04  & 31.39\footnotesize$\pm$\footnotesize 0.05  & 39.55\footnotesize$\pm$\footnotesize 0.12  & 36.38\footnotesize$\pm$\footnotesize 0.07  & 32.85 \\
& \multirow{2}{*}{\ding{52}} & \multirow{2}{*}{-} & - & 33.68\footnotesize$\pm$\footnotesize 0.03  & 27.62\footnotesize$\pm$\footnotesize 0.04  & 28.81\footnotesize$\pm$\footnotesize 0.07  & 31.42\footnotesize$\pm$\footnotesize 0.07  & 39.85\footnotesize$\pm$\footnotesize 0.13  & 36.71\footnotesize$\pm$\footnotesize 0.09  & 33.02 \\
&  &  & \textbf{5 B} & 34.61\footnotesize$\pm$\footnotesize 0.08  & 28.46\footnotesize$\pm$\footnotesize 0.06  & 29.72\footnotesize$\pm$\footnotesize 0.10  & 32.29\footnotesize$\pm$\footnotesize 0.03  & 40.64\footnotesize$\pm$\footnotesize 0.07  & 37.48\footnotesize$\pm$\footnotesize 0.05  & 33.87 \\
& \multirow{2}{*}{\ding{52}} & \multirow{2}{*}{\ding{52}} & - & 34.23\footnotesize$\pm$\footnotesize 0.06  & 27.96\footnotesize$\pm$\footnotesize 0.03  & 29.26\footnotesize$\pm$\footnotesize 0.11  & 31.93\footnotesize$\pm$\footnotesize 0.08  & 40.16\footnotesize$\pm$\footnotesize 0.03  & 37.21\footnotesize$\pm$\footnotesize 0.04  & 33.46 \\
&  &  & \textbf{5 B} & 35.10\footnotesize$\pm$\footnotesize 0.04  & 28.73\footnotesize$\pm$\footnotesize 0.04  & 29.97\footnotesize$\pm$\footnotesize 0.07  & 32.64\footnotesize$\pm$\footnotesize 0.12  & 40.91\footnotesize$\pm$\footnotesize 0.06  & 37.93\footnotesize$\pm$\footnotesize 0.10  & 34.21 \\
\hline
\multirow{6}{*}{\textbf{E-ARM}} & \multirow{2}{*}{-} & \multirow{2}{*}{-} & - & 32.99\footnotesize$\pm$\footnotesize 0.10  & 27.15\footnotesize$\pm$\footnotesize 0.03  & 28.33\footnotesize$\pm$\footnotesize 0.12  & 31.13\footnotesize$\pm$\footnotesize 0.04  & 39.56\footnotesize$\pm$\footnotesize 0.01  & 36.07\footnotesize$\pm$\footnotesize 0.02  & 32.54 \\
&  &  & \textbf{5 B} & 34.06\footnotesize$\pm$\footnotesize 0.06  & 27.97\footnotesize$\pm$\footnotesize 0.08  & 29.26\footnotesize$\pm$\footnotesize 0.09  & 31.90 \footnotesize$\pm$\footnotesize 0.13 & 40.30 \footnotesize$\pm$\footnotesize 0.03 & 36.92 \footnotesize$\pm$\footnotesize 0.09 & 33.40 \\
& \multirow{2}{*}{\ding{52}} & \multirow{2}{*}{-} & - & 33.97 \footnotesize$\pm$\footnotesize 0.08 & 28.03 \footnotesize$\pm$\footnotesize 0.04 & 29.13 \footnotesize$\pm$\footnotesize 0.02 & 31.84 \footnotesize$\pm$\footnotesize 0.11 & 40.32 \footnotesize$\pm$\footnotesize 0.03 & 36.96 \footnotesize$\pm$\footnotesize 0.07 & 33.38 \\
&  &  & \textbf{5 B} & 34.93 \footnotesize$\pm$\footnotesize 0.05 & 28.91 \footnotesize$\pm$\footnotesize 0.12 & 30.04 \footnotesize$\pm$\footnotesize 0.11 & 32.56 \footnotesize$\pm$\footnotesize 0.04 & 41.01 \footnotesize$\pm$\footnotesize 0.06 & 37.73 \footnotesize$\pm$\footnotesize 0.12 & 34.20 \\
& \multirow{2}{*}{\ding{52}} & \multirow{2}{*}{\ding{52}} & - & 34.58 \footnotesize$\pm$\footnotesize 0.09 & 28.38 \footnotesize$\pm$\footnotesize 0.12 & 29.56 \footnotesize$\pm$\footnotesize 0.10 & 32.11 \footnotesize$\pm$\footnotesize 0.03 & 40.93 \footnotesize$\pm$\footnotesize 0.03 & 37.56 \footnotesize$\pm$\footnotesize 0.07 & 33.85 \\
&  &  & \textbf{5 B} & \textbf{35.36} \footnotesize$\pm$\footnotesize 0.05 & \textbf{29.11} \footnotesize$\pm$\footnotesize 0.04 & \textbf{30.25} \footnotesize$\pm$\footnotesize 0.09 & \textbf{32.82} \footnotesize$\pm$\footnotesize 0.11 & \textbf{41.58} \footnotesize$\pm$\footnotesize 0.07 & \textbf{38.19} \footnotesize$\pm$\footnotesize 0.03 & \textbf{34.55} \\
\bottomrule
\end{tabular}%
}
\caption{\footnotesize{Comparison of BLEU scores between our approach E-ARM and the base  ARGM trained just with cross-entropy loss on six translation pairs of IWSLT14 datasets. We use ``-'' to denote that the training trick is not used while ``\ding{52}'' indicates we use it. ``\textbf{5 B}'' represents we use beam searching with 5 beams.}}

\label{tab:iwslt14-6PAIRS}
\end{table*}
E-ARM is first evaluated in the context of neural machine translation (NMT), which is a conditional generation task and is important in the natural language processing (NLP) field. We first analyze E-ARM on the IWSLT14 dataset, which includes six different language pairs (\{German, Spanish, Italian\} $\rightarrow$ English and English $\rightarrow$ \{German, Spanish, Italian\}). In addition,  we test E-ARM on the WMT16 (English $\rightarrow$ German) benchmark to make sure we evaluating E-ARM on a larger dataset. Hereafter we abbreviate English, German, Spanish, Italian as ``En'', ``De'', ``Es'', ``It''. We use one size of transformer (``Base-IWSLT'') for the IWSLT14 benchmark and two sizes of transformer (``Base-WMT'', ``Large-WMT'') for the WMT16 benchmark. Scheduled Sampling is carried out following~\citet{SSTransformerMihaylovaM19}. More experimental details are reported in Appendix ~\ref{sec:expr_setting}.

The results of IWSLT14 tasks are shown in Table \ref{tab:iwslt14-6PAIRS}. We test not only the pure performance of E-ARM but also the compatibility with other techniques. In detail, we can observe that (1) without any particular engineering, E-ARM outperforms the base autoregressive translation model trained with cross-entropy singly by 0.62 (31.92 $\rightarrow$ 32.54) BLEU points in average, especially on three translation pairs---38.61 $\rightarrow$ 39.56 on Spanish-to-English, 30.48 $\rightarrow$ 31.13 on Italian-to-English, 35.42 $\rightarrow$ 36.07 on English-to-Spanish. (2) E-ARM is compatible with other  techniques like scheduled sampling, which can help alleviate the exposure bias problem to some extent. They are not mutually exclusive and can work together to further improve the performance of the base ARGM. (3) However, since scheduled sampling can reduce exposure bias and beam search can somewhat alleviate the flaws caused by greedy selection at each time step, the performance gain of E-ARM when all these tactics are combined is only 0.34 (34.21 $\rightarrow$ 34.55), which is lower than the 0.62 (31.92 $\rightarrow$ 32.54) obtained when the model is purely trained without these other techniques.
\begin{table}[htp]
\centering
\renewcommand{\arraystretch}{1.2}
\resizebox{0.8\linewidth}{!}{
\begin{tabular}{c|ccc|c}
\toprule
\textbf{Model} & \textbf{L.S.} & \textbf{S.S.} & \textbf{w/E-ARM} & \textbf{BLEU \bm$\uparrow$} \\
\midrule[1.2pt]
\multirow{4}{*}{ \textbf{Base-WMT}} & - & - & - & 27.56 \\
 & \ding{52} & - & - & 28.04  \\
 & \ding{52} & \ding{52} & - & 28.36  \\
 & \ding{52} & \ding{52} & \ding{52} & \textbf{28.62} \\
\hline
\multirow{4}{*}{ \textbf{Large-WMT}} & - & - & - & 28.70  \\
 & \ding{52} & - & - & 29.05  \\
 & \ding{52} & \ding{52} & - & 29.23  \\
 & \ding{52} & \ding{52} & \ding{52} & \textbf{29.44} \\
\bottomrule
\end{tabular}
}
\tabcaption{\footnotesize{Translation performance of proposed E-ARM on WMT16 English$\rightarrow$German, evaluated with BLEU. We uniformly use 5 beams when applying beam search. ``\textbf{L.S.}'' denotes Label Smoothing and ``\textbf{S.S.}'' denotes Scheduled Sampling.}}
\label{tab:wmt16}
\end{table}

Additionally, Table \ref{tab:wmt16} shows the performance of E-ARM on the WMT16 English $\rightarrow$ German task. For two different model sizes, enabling label smoothing (L.S.) improves model performance by 0.52 and 0.35, respectively. The performance of the base transformer model further increases to 28.36 BLEU points when scheduled sampling (S.S.) is used, while the larger model improves to 29.23 points. E-ARM paired with label smoothing and scheduled sampling yields the highest scores of 28.62 and 29.44, respectively. Overall, our training strategy outperforms ARGM's vanilla teacher-forcing training and can have uniformly favorable impacts across different models and dataset sizes.

\subsection{Application to Language Modeling}
\label{sec:app_lm}
\begin{table*}[htp]
\centering
\renewcommand{\arraystretch}{1.2}
\resizebox{0.8\linewidth}{!}{
\begin{tabular}{l|c|ccc}
\toprule
\multirow{2}{*}{\textbf{Model}} & \textbf{Energy} & \multicolumn{3}{c}{\textbf{PPL \bm$\downarrow$}} \\ 
\cline{3-5}
& \textbf{Re-sampling} & \textbf{CC-News} & \textbf{Toronto Book Corpus} & \textbf{WikiText103}  \\
\midrule[1.2pt]
\textbf{Tr-Base} & - &  18.29 & 17.57 & 30.56\\
\textbf{Residual EBM(Tr-Base)} & \ding{52}& \textbf{15.57-15.58} & 16.98-17.00  & 29.88-29.93 \\
\textbf{Tr-XL} & -  & - & - & 24.20 \\
\textbf{Residual EBM(Tr-XL)} & \ding{52}  & - & - & 23.85-23.87 \\
\hline
\textbf{E-ARM(Tr-Base)}  & - & 15.78 & 17.10 & 29.94 \\
\textbf{E-ARM(Tr-Base)}  & \ding{52} & 15.63-15.67 & \textbf{16.89-16.93} &  29.81-29.84\\
\textbf{E-ARM(Tr-XL)} & -  & - & - & 23.90 \\
\textbf{E-ARM(Tr-XL)} & \ding{52}  & - & - & \textbf{23.79-23.82} \\
\bottomrule
\end{tabular}
}
\tabcaption{\footnotesize{Language modeling performance of different models on three benchmarks. Evaluation is conducted using perplexity (PPL). We test the performance of E-ARM w/o energy resampling technique. The residual EBM~\cite{ResidualBakhtinDGORS21}} requires an extra model having the same structure with the underlying autoregressive models to learn the energy so that doubles parameters.}
\label{tab:language_modeling}
\end{table*}

To further demonstrate E-ARM's consistency in reducing flaws of autoregressive generative models, we also conduct experiments on language modeling tasks. Three different datasets, WikiText-103~\citep{Wikitext103}, Toronto Book Corpus~\cite{bookcorpus1, bookcorpus2}, and CC-news~\cite{cc-news}, are chosen as the testbed.  WikiText-103 comprises 103 million training tokens from 28 thousand articles, with an average length of 3.6 thousand tokens per article; Toronto Book Corpus consists of fiction books in 16 different genres, totaling about half a billion words; and CC-news is a de-duplicated subset of the English portion
of the CommonCrawl news dataset, which totals around 16 Billion words. Two autoregressive network structures are used to evaluate our method's effectiveness: vanilla Transformer~\citep{TransformerBase} (``Tr-Base'' for short) tested on all three benchmarks and  Transformer-XL~\citep{TransformerXL} (``Tr-XL'' for short), which is a transformer equipped with a recurrent memory, tested on Wikitext-103. In particular, residual EBM~\cite{residualDengBOSR20} shares a similar energy-based idea to improve the quality of text generation for autoregressive models, though they treat the underlying autoregressive model and the EBM as two independent models and apply the distribution modeling on the entire sequence instead of subsequences with respect to each time step. As a result, we implement it as a baseline of our method. Besides, the Top-K energy resampling post-processing technique, which is a critical module of residual EBM, is also applicable for our E-ARM since we can estimate the energy of entire input sequences\footnote{It worth noting that Top-K energy resampling can not get the PPL directly. ~\citet{ResidualBakhtinDGORS21} provides a way to approximate PPL, which leads to an estimated interval of PPL. }.


The final results are reported in Table~\ref{tab:language_modeling}. We can see from the results that E-ARM outperforms two pure autoregressive models with clear margins over all three benchmarks. Specifically, on Wikitext-103 benchmark, our E-ARM improves the performance of Transformer-Base model and Transformer-XL model by 0.62 PPL points (from 30.56 to 29.94) and 0.30 PPL points (from 24.20 to 23.90) respectively; on CC-news and Toronto Book Corpus benchmarks, our method obtains 0.51 ppl and 0.47 ppl performance gain respectively, and gets further improvement once energy resampling technique was applied. Besides, though residual EBM's learning parameters are twice as ours and their method is unable to directly benefit autoregressive models without energy resampling, our E-ARM achieves comparable results to them, even slightly better on Toronto Book Corpus and Wikitext-103 benchmarks.

\begin{table}[htp]
\centering
\renewcommand{\arraystretch}{1.2}
\resizebox{0.9\linewidth}{!}{
\begin{tabular}{c|ccccc}
\toprule
\textbf{Model} &  \multicolumn{5}{c}{\textbf{Start Epoch of E-ARM}} \\
\cline{2-6}
\textbf{Structure} & \textbf{5} & \textbf{10} & \textbf{15} & \textbf{20} & \textbf{25}  \\
\hline
\textbf{Tr-Base} & 30.38 & 30.12 & \textbf{29.94} & 30.05 & 30.29 \\
\textbf{Tr-XL} & 24.12 & \textbf{23.90} & 23.96 & 24.05 & 24.16 \\
\bottomrule
\end{tabular}
}
\tabcaption{\footnotesize{Exploring the effect of different start epochs of E-ARM on Wikitext103 benchmark. Performances are evaluated by perplexity (PPL).}}
\label{tab:ablation_lm}
\end{table}

In addition, we have studied the effect of different start epochs of E-ARM on the performance of language modeling, which can be seen in Table~\ref{tab:ablation_lm}. From this, we may deduce that starting E-ARM training at the 15-th and 10-th epoch yields the best results for Transformer-Base and Transformer-XL respectively, whereas starting earlier or later yields a performance decline. It is reasonable because, if E-ARM was introduced too early, the autoregressive model may not have been optimized well at that moment. As a result, the quality of generation for ``negative phase'' would be terrible, making energy-based training unstable. On the other hand, the underlying autoregressive model can be modified only marginally if E-ARM was introduced when the ARGM training is virtually complete. 

\subsection{Application to Image Generation}
In order to illustrate the effectiveness and generality of our method in processing different modality tasks, we further show the results of applying E-ARM to image generation in this section. We apply E-ARM to Pixel-CNN~\citep{van2016pixel} and its variant Gated Pixel-CNN~\citep{oord2016conditional}. Experiments are carried out on the MNIST and CIFAR-10 datasets.

\begin{table}[htp]
\centering
\renewcommand{\arraystretch}{1.2}
\resizebox{0.9\linewidth}{!}{
\begin{tabular}{l|cc}
\toprule
\multirow{2}{*}{\large\textbf{Model}} & \multicolumn{2}{c}{\textbf{Test (Train) NLL \bm$\downarrow$}} \\ 
\cline{2-3}
& \textbf{MNIST} & \textbf{CIFAR-10} \\
\midrule[1.2pt]
\textbf{Pixel-CNN}         & 0.17 (0.13) & 3.14 (3.08) \\
\textbf{Pixel-CNN (w/E-ARM)} & \textbf{0.15 (0.12)} & \textbf{3.07 (2.98)} \\
\textbf{Gated Pixel-CNN}   & 0.14 (0.11) & 3.03 (2.90) \\
\textbf{Gated Pixel-CNN (w/E-ARM)} & \textbf{0.12 (0.10)} & \textbf{2.97 (2.87)}  \\
\bottomrule
\end{tabular}
}
\\
\tabcaption{Performance of E-ARM with different base networks on MNIST and CIFAR-10 in bits/dim (lower is better), training performance in brackets.}
\label{tab:image_generation}
\end{table} 

Table~\ref{tab:image_generation} summarizes the quantitative results measured by per-pixel negative log-likelihood (NLL). 
We can see that with the help of our E-ARM, both the Pixel-CNN and the Gated Pixel-CNN can obtain improvements in all datasets (0.17 $\rightarrow$ 0.15 and 3.14 $\rightarrow$ 3.07 for Pixel-CNN on MNIST and CIFAR10 respectively and 0.14 $\rightarrow$ 0.12 and 3.03 $\rightarrow$ 2.97 for Gated Pixel-CNN on MNIST and CIFAR10 respectively). This is further evidence in favour of the energy-based learning objective for improving autoregressive models. 

\section{Related Works}
\label{sec:related_work}
\subsection{Autoregressive Generative Models}
Modeling high-dimensional data distributions directly is usually a rather challenging task due to ``the curse of dimensionality''~\citep{DynamicProgramming}. One alternative method is to sequentialize the random variables and then factorize the joint probability distribution into the product of conditionals based on the sequence structure, which is exactly the core idea of autoregressive generative models (ARGMs). ARGMs have been very successful, in particular for sequential data. For example, ARGMs have been widely used in language modeling~\citep{TransformerBase, TransformerXL, GPT2}, audio synthesis~\citep{wavenet}, and even image generation~\citep{pixelrnn, pixelcnnOordKEKVG16, pixelcnn++SalimansK0K17}. The advantages of ARGMs are however balanced by issues of (1) exposure bias~\citep{SSRanzatoCAZ15, SSBengioVJS15, SongT020}, due to the discrepancy in input context distributions between the training and inference stages, and (2) weak long-range coherence, due to the inherent greedy selection of one token at a time without look-ahead. Recently, instead of constructing an autoregressive model in the data space, \citet{AnytimeSampling} have proposed a unique way which uses autoregressive models in the latent space followed by a decoder which decodes the autoregressively generated latent feature into the original data space, trading off the sample quality for computational efficiency.

\subsection{Energy-Based Models}

The seminal idea of combing a generative model and an energy-based model has been explored by a plethora of great works~\citep{LearningLatentSpaceEBPM, autoregressiveEnergyDurkanN19, CooperativeTrainingFASTSLOW, CooperativeTrainingDescGen, VAEBMXiaoKKV21, ResidualBakhtinDGORS21, YourGanCheZSLPCB20, arbel2021generalized}. In particular, \citet{LearningLatentSpaceEBPM} aimed to learn an energy-based model (EBM) in the latent space of a generator model, so that the EBM can act as a prior model on the generator model's top-down network. VAEBM, a symbiotic composition of a variational auto-encoder and an EBM, was proposed by \citep{VAEBMXiaoKKV21}.  ~\citet{arbel2021generalized} proposed a novel training method for the combined model of GANs and EBMs by leveraging the Donsker-Varadham representation of KL-divergence. 


Besides, due to the challenge of sampling from EBMs, training EBMs by contrastive divergence methods~\citep{Hinton02Training, DeepDirectedKimB16, grathwohl2021no} is difficult, especially on high-dimensional data. MCMC methods~\citep{nijkamp2019learning,du2019implicit,jemGrathwohlWJD0S20} are usually adopted for data sampling. However, these methods require enormous extra computing overheads and are not applicable when the input is discrete such as text sequences~\citep{residualDengBOSR20}. As a result, a variety of recent works attempt to explore the strategy of training an EBM without MCMC. In particular, ~\citet{ResidualBakhtinDGORS21,  MCGminkaixu, FLOWNCE} optimize the EBMs by using noise contrastive estimation (NCE)~\citep{NCE-EBM, ConditionalNCE-EBM}. ~\citet{autoregressiveEnergyDurkanN19} estimate the intractable normalization component by utilizing ARGMs and importance sampling. ~\citet{BetterMixing, YourGanCheZSLPCB20, EOWSoftmax} skirt the challenge of collecting data in the high-dimensional data space by performing sampling using a carfefully crafted latent space, which improves sampling efficiency. 

\section{Conclusions and Future Work}
In this paper, we propose a novel training method dubbed E-ARM for autoregressive models by treating them as an energy-based model. This is achieved by defining the energy function using the softmax operation's extra degree of freedom within an autoregressive network. We further design a unique way to improve the training of E-ARM using importance sampling. Experimental results on two language tasks and one vision task demonstrate the effectiveness of E-ARM to alleviate exposure bias and incoherence problems of ARGMs. In the future, we expect to extend E-ARM on other sequential generation tasks (\textit{e.g.} text summarization, audio generation), and incorporate the proposed methodology into other advanced autoregressive architectures.

\clearpage
\bibliography{example_paper}
\bibliographystyle{icml2022}

\clearpage
\appendix
\onecolumn
\section{Derivation of the Negative Phase Gradient}
\label{appendix:A}
In this section, we show the detailed derivation of Eq. \ref{eq:sleep_phase_derived}. Formally, as shown in Sec. \ref{sec:model}, given an autoregressive model $q_\theta(\mathbf{x}_{< k})=\prod_{l=1}^{k-1}q_\theta(\mathbf{x}_l|\mathbf{x}_{< l})$ ($k$ denotes the time step) with parameters $\theta$, we define a product of the autoregressive model and an EBM as follows
\begin{equation}
\begin{aligned}
\label{eq:prod_ebm_app}
p_\theta(\mathbf{x}_k, \mathbf{x}_{< k}) = q_\theta(\mathbf{x}_{< k}) \cdot \frac{e^{-\phi_{\theta}(\mathbf{x}_k, \mathbf{x}_{< k})}}{\mathbf{Z}_\theta},
\end{aligned}
\end{equation}
where $q_\theta(\mathbf{x}_{< k}) = \prod_{l=1}^{k-1}q_\theta(\mathbf{x}_l|\mathbf{x}_{< l})$, $\mathbf{Z}_\theta$ is the normalization term and equal to $\mathbb{E}_{\mathbf{x}'_{< k} \sim Q_\theta(\mathbf{x}_{< k})}[ \sum_{\mathbf{x}_k}  e^{-\phi_{\theta}(\mathbf{x}_k, \mathbf{x}'_{< k})}]$. The optimization of $p_\theta(\mathbf{x}_k, \mathbf{x}_{< k})$ includes two phases, and the gradient w.r.t $\theta$ of ``negative phase'' is
\begin{equation}
\begin{aligned}
\label{eq:sleep_phase_app}
-\mathbb{E}_{\mathbf{x}_{< k} \sim P_\theta}[\nabla_{\theta}\log q_\theta (\mathbf{x}_{< k}) ] + \mathbb{E}_{\mathbf{x}_k, \mathbf{x}_{< k}\sim P_\theta}[\nabla_{\theta}\phi_{\theta}(\mathbf{x}_k, \mathbf{x}_{< k})].
\end{aligned}    
\end{equation}
Next, we will show the specific derivation about how to transform Eq. \ref{eq:sleep_phase_app} into Eq. \ref{eq:sleep_phase_derived}.

\subsection{Derivation of the First Term}
The first term $\mathbb{E}_{\mathbf{x}_{< k} \sim P_\theta}[\nabla_{\theta}\log q_\theta (\mathbf{x}_{< k}) ]$ can be processed as follows
\begin{equation}
\begin{aligned}
\mathbb{E}_{\mathbf{x}_{< k} \sim P_\theta}[\nabla_{\theta}\log q_\theta (\mathbf{x}_{< k}) ] =& \sum_{\mathbf{x}_{< k}} p_\theta(\mathbf{x}_{< k}) \nabla_{\theta}\log q_\theta (\mathbf{x}_{< k}) \\ =& \sum_{\mathbf{x}_{< k}} \sum_{\mathbf{x}_k} p_\theta(\mathbf{x}_k, \mathbf{x}_{< k})  \nabla_{\theta}\log q_\theta (\mathbf{x}_{< k}) \\ =&\sum_{\mathbf{x}_{< k}} q_\theta (\mathbf{x}_{< k})  \frac{\sum_{\mathbf{x}_k}e^{-\phi_\theta(\mathbf{x}_k, \mathbf{x}_{< k})}}{\mathbf{Z}_\theta} \nabla_{\theta}\log q_\theta (\mathbf{x}_{< k})\\=&\mathbb{E}_{\mathbf{x}_{< k} \sim Q_\theta (\mathbf{x}_{< k}) }[\mathbf{w}(\mathbf{x}_{< k})\nabla_{\theta} \log q_\theta (\mathbf{x}_{< k})  ]\label{eq:A1_4},
\end{aligned}    
\end{equation}
where we have $\mathbf{w}(\mathbf{x}_{< k}) =\frac{\sum_{\mathbf{x}_k}e^{-\phi(\mathbf{x}_k, \mathbf{x}_{< k})}}{\mathbb{E}_{\mathbf{x}'_{< k} \sim Q_\theta(\mathbf{x}_{< k})}[ \sum_{\mathbf{x}_k}  e^{-\phi_{\theta}(\mathbf{x}_k, \mathbf{x}'_{< k})}]}$ because
\begin{equation}
\begin{aligned}
\mathbf{w}(\mathbf{x}_{< k})&=\frac{\sum_{\mathbf{x}_k}e^{-\phi(\mathbf{x}_k, \mathbf{x}_{< k})}}{\mathbf{Z}_\theta} = \frac{\sum_{\mathbf{x}_k}e^{-\phi(\mathbf{x}_k, \mathbf{x}_{< k})}}{\sum_{\mathbf{x}_{< k}} \sum_{\mathbf{x}_k} q_\theta(\mathbf{x}_{< k}) e^{-\phi_{\theta}(\mathbf{x}_k, \mathbf{x}_{< k})}} \\ &=\frac{\sum_{\mathbf{x}_k}e^{-\phi(\mathbf{x}_k, \mathbf{x}_{< k})}}{\sum_{\mathbf{x}_{< k}} q_\theta(\mathbf{x}_{< k}) \sum_{\mathbf{x}_k}  e^{-\phi_{\theta}(\mathbf{x}_k, \mathbf{x}_{< k})}} \\&=\frac{\sum_{\mathbf{x}_k}e^{-\phi(\mathbf{x}_k, \mathbf{x}_{< k})}}{\mathbb{E}_{\mathbf{x}_{< k} \sim Q_\theta(\mathbf{x}_{< k})}[ \sum_{\mathbf{x}_k}  e^{-\phi_{\theta}(\mathbf{x}_k, \mathbf{x}_{< k})}]}. 
\end{aligned}    
\end{equation}

\subsection{Derivation of the Second Term}
Then, we tackle the second term $\mathbb{E}_{\mathbf{x}_k, \mathbf{x}_{< k}\sim P_\theta}[\nabla_{\theta}\phi_{\theta}(\mathbf{x}_k, \mathbf{x}_{< k})]$ as follows
\begin{equation}
\begin{aligned}
\mathbb{E}_{P_\theta}\big[ \nabla_{\theta} \phi_\theta(\mathbf{x}_k, \mathbf{x}_{< k}) \big] &= \sum_{\mathbf{x}_k, \mathbf{x}_{< k}} p_\theta(\mathbf{x}_k, \mathbf{x}_{< k})\nabla_{\theta} \phi_\theta(\mathbf{x}_k, \mathbf{x}_{< k}) \\&= \sum_{\mathbf{x}_k, \mathbf{x}_{< k}} p_\theta(\mathbf{x}_k, \mathbf{x}_{< k}) \frac{q_\theta(\mathbf{x}_k, \mathbf{x}_{< k})}{q_\theta(\mathbf{x}_k, \mathbf{x}_{< k})}\nabla_{\theta}\phi_\theta(\mathbf{x}_k, \mathbf{x}_{< k}) \\&=\sum_{\mathbf{x}_k, \mathbf{x}_{< k}} q_\theta(\mathbf{x}_k, \mathbf{x}_{< k}) \frac{q_\theta(\mathbf{x}_{< k}) \cdot e^{-\phi_{\theta}(\mathbf{x}_k, \mathbf{x}_{< k})}}{ \mathbf{Z}_\theta \cdot q_\theta(\mathbf{x}_k, \mathbf{x}_{< k})}\nabla_{\theta}\phi_\theta(\mathbf{x}_k, \mathbf{x}_{< k})
\\&=\mathbb{E}_{\mathbf{x}_k, \mathbf{x}_{< k}\sim Q_\theta(\mathbf{x}_k, \mathbf{x}_{< k})}[\frac{e^{-\phi_{\theta}(\mathbf{x}_k, \mathbf{x}_{< k})}}{q_\theta(\mathbf{x}_k | \mathbf{x}_{< k})}\cdot \frac{1}{\mathbf{Z}_\theta}\nabla_{\theta}\phi_\theta(\mathbf{x}_k, \mathbf{x}_{< k})]\\&=\sum_{\mathbf{x}_{< k}} q_\theta (\mathbf{x}_{< k})  \sum_{\mathbf{x}_k} q_\theta(\mathbf{x}_k | \mathbf{x}_{< k})\frac{e^{-\phi_{\theta}(\mathbf{x}_k, \mathbf{x}_{< k})}}{q_\theta(\mathbf{x}_k | \mathbf{x}_{< k})}\cdot \frac{1}{\mathbf{Z}_\theta}\nabla_{\theta}\phi_\theta(\mathbf{x}_k, \mathbf{x}_{< k})\\&=\sum_{\mathbf{x}_{< k}} q_\theta (\mathbf{x}_{< k}) \sum_{\mathbf{x}_k} e^{-\phi_{\theta}(\mathbf{x}_k, \mathbf{x}_{< k})} \cdot \frac{1}{\mathbf{Z}_\theta}\nabla_{\theta}\phi_\theta(\mathbf{x}_k, \mathbf{x}_{< k}) \\&= \mathbb{E}_{Q_\theta (\mathbf{x}_{< k}) }[\sum_{\mathbf{x}_k} \frac{e^{-\phi_{\theta}(\mathbf{x}_k, \mathbf{x}_{< k})} }{\mathbf{Z}_\theta}\nabla_{\theta}\phi_\theta(\mathbf{x}_k, \mathbf{x}_{< k}) ] \\
&=\mathbb{E}_{Q_\theta (\mathbf{x}_{< k}) }[\sum_{\mathbf{x}_k}\frac{e^{-\phi_{\theta}(\mathbf{x}_k, \mathbf{x}_{< k})}}{\sum_{\mathbf{x}_k}e^{-\phi_{\theta}(\mathbf{x}_k, \mathbf{x}_{< k})}}\cdot\frac{\sum_{\mathbf{x}_k}e^{-\phi_{\theta}(\mathbf{x}_k, \mathbf{x}_{< k})}}{\mathbf{Z}_\theta}\nabla_{\theta}\phi_\theta(\mathbf{x}_k, \mathbf{x}_{< k})]\\&=\mathbb{E}_{Q_\theta (\mathbf{x}_{< k}) }[\sum_{\mathbf{x}_k}q_\theta(\mathbf{x}_k | \mathbf{x}_{< k})\mathbf{w}(\mathbf{x}_{< k})\nabla_{\theta}\phi_\theta(\mathbf{x}_k, \mathbf{x}_{< k})]\\&=\mathbb{E}_{Q_\theta (\mathbf{x}_{< k}) }[\mathbb{E}_{a \sim Q_\theta(\mathbf{x}_k | \mathbf{x}_{< k})}[\mathbf{w}(\mathbf{x}_{< k})\nabla_{\theta}\phi_\theta(\mathbf{x}_k, \mathbf{x}_{< k})]]\\&=\mathbb{E}_{\mathbf{x}_k, \mathbf{x}_{< k}\sim Q_\theta(\mathbf{x}_k, \mathbf{x}_{< k})}[\mathbf{w}(\mathbf{x}_{< k})\nabla_{\theta}\phi_\theta(\mathbf{x}_k, \mathbf{x}_{< k})]\label{eq:A2_10}
\end{aligned}    
\end{equation}
where $\mathbf{w}(\mathbf{x}_{< k})$ is also equal to $\frac{\sum_{\mathbf{x}_k}e^{-\phi(\mathbf{x}_k, \mathbf{x}_{< k})}}{\mathbf{Z}_\theta}$. Combining Eq. \ref{eq:A1_4} and Eq. \ref{eq:A2_10}, we can obtain an equivalent form of the gradient of the negative phase  without any  expectation over $p_\theta$ as 
\begin{gather}
-\mathbb{E}_{\mathbf{x}_{< k} \sim Q_\theta (\mathbf{x}_{< k}) }[\mathbf{w}(\mathbf{x}_{< k})\nabla_{\theta} \log q_\theta (\mathbf{x}_{< k})  ] + \mathbb{E}_{\mathbf{x}_k, \mathbf{x}_{< k}\sim Q_\theta(\mathbf{x}_k, \mathbf{x}_{< k})}[\mathbf{w}(\mathbf{x}_{< k})\nabla_{\theta}\phi_\theta(\mathbf{x}_k, \mathbf{x}_{< k})], \\
\textbf{where} \quad \mathbf{w}(\mathbf{x}_{< k})  =  \frac{\sum_{\mathbf{x}_k}e^{-\phi(\mathbf{x}_k, \mathbf{x}_{< k})}}{\mathbb{E}_{\mathbf{x}'_{< k} \sim Q_\theta(\mathbf{x}_{< k})}[ \sum_{\mathbf{x}_k}  e^{-\phi_{\theta}(\mathbf{x}_k, \mathbf{x}'_{< k})}]}.
\end{gather}

\section{Further Refinement of $\mathbf{w}$}
The reweighing weight $\mathbf{w}$ can be further deduced as
\label{sec:appendix_w}
\begin{equation}
\begin{aligned}
\mathbf{w}(\mathbf{x}_{< k})  &=  \frac{\sum_{\mathbf{x}_k}e^{-\phi(\mathbf{x}_k, \mathbf{x}_{< k})}}{\mathbb{E}_{\mathbf{x}'_{< k} \sim Q_\theta(\mathbf{x}_{< k})}[ \sum_{\mathbf{x}_k}  e^{-\phi_{\theta}(\mathbf{x}_k, \mathbf{x}'_{< k})}]} = \frac{\sum_{\mathbf{x}_k}\frac{p_\theta(\mathbf{x}_k, \mathbf{x}_{< k})}{q_\theta(\mathbf{x}_{< k})}}{\mathbb{E}_{\mathbf{x}'_{< k} \sim Q_\theta(\mathbf{x}_{< k})}[ \sum_{\mathbf{x}_k}  \frac{p_\theta(\mathbf{x}_k, \mathbf{x}_{< k})}{q_\theta(\mathbf{x}_{< k})}]} \\ &= \frac{\frac{p_\theta(\mathbf{x}_{< k})}{q_\theta(\mathbf{x}_{< k})}}{\mathbb{E}_{\mathbf{x}'_{< k} \sim Q_\theta(\mathbf{x}_{< k})}[ \frac{p_\theta(\mathbf{x}_{< k})}{q_\theta(\mathbf{x}_{< k})}]} =\frac{\mu(\mathbf{x}_{< k})}{\mathbb{E}_{\mathbf{x}'_{< k}} \mu(\mathbf{x}_{< k})},
\end{aligned}   
\end{equation}
where $\mu(\mathbf{x}_{< k})$ is defined as $\frac{p_\theta(\mathbf{x}_{< k})}{\tilde{q}_\theta(\mathbf{x}_{< k})}$.

\section{Experimental Settings}
\label{sec:expr_setting}
In this section, we introduce the specific setup of different benchmarks in Table \ref{tab:hyper-parameter}. We uniformly use Adam optimizer. The training will be stopped once the model has not obtained better performance for 20 epochs on the validation set. For translation tasks, the length of generated fake sentences, which is used for the computing of negative phase in Eq.~\ref{eq:final_objective}, is dependent on the source sequence whilst for language modeling tasks, we fix the length of generated fake sentences as 50 during training.  As for the model structures of the image generation task, we use the official structure reported by PixelCNN~\citep{pixelrnn} and Gated PixelCNN~\citep{pixelcnnOordKEKVG16} without modification. We use the same batch of samples generated autoregressively to approximate both the expectations in Eq.\ref{eq:final_objective} and weight \textbf{w} (\textit{i.e.}, shared), which does not need to sample twice. The number of samples in a batch is dynamic while the maximum number of the total tokens in a batch are fixed (4096 in our experiments). If the length of sequences in a batch is 32, then it includes 4096 / 32 = 128 samples in total. It is a common strategy in language generation tasks, and has been used in many frameworks(e.g. Fairseq~\citep{ott2019fairseq}). We generate samples autoregressively as many as the number of sequences in the current batch at each update iteration. 
\begin{table}[h]
\centering
\renewcommand{\arraystretch}{1.2}
\resizebox{0.75\linewidth}{!}{
\begin{tabular}{l|c|cc|cc}
\toprule
\multirow{2}{*}{\centering\textbf{Hyper-Parameters}}  & \textbf{Translation-IWSLT14} & \multicolumn{2}{c|}{\textbf{Translation-WMT16}} & \multicolumn{2}{c}{\textbf{Language Modeling}} \\
\cline{2-6}
& \textbf{Tr-Base} & \textbf{Tr-Base} & \textbf{Tr-Large} & \textbf{Tr-Base} & \textbf{Tr-XL} \\
\midrule[1.2pt]
\textbf{Number of Layers} & 12 & 12 & 12 & 6 & 16 \\
\textbf{Hidden Embed Size} & 512 & 512 & 1024 & 512 & 410 \\
\textbf{FC-Layer Embed Size} & 1024 & 2048 & 4096 & 2048 & 2100 \\
\textbf{Attention Heads} & 4 & 8 & 16 & 8 & 10 \\
\textbf{Dropout} & 0.3 & 0.3 & 0.3 & 0.1 & 0.1 \\
\textbf{Learning Rate} & 5e-4 & 1e-3 & 1e-3 & 5e-4 & 2.5e-4  \\
\textbf{lr scheduler} & inverse\_sqrt & inverse\_sqrt & inverse\_sqrt & inverse\_sqrt & cosine  \\
\textbf{Warm up Updates} & 4000 & 4000 & 4000 & 4000 & 10000  \\
\textbf{Weigth Decay} & 1e-4 & 0.0 & 0.0 & 1e-2 &  0.0 \\
\textbf{E-ARM Start Epoch} & 15 & 15 & 10 & 15 & 10  \\
\bottomrule
\end{tabular}
}
\tabcaption{\footnotesize{Hyper-Parameters of different model structures and datasets. ``Tr-Base'', ``Tr-Large'', and ``Tr-XL'' indicate Transformer-Base, Transformer-Large, and Transformer-XL respectively}}
\label{tab:hyper-parameter}
\end{table}


\section{More Experimental Analysis}
\label{sec:more_experiment}
\subsection{Effect on Incoherence}

In order to validate the effectiveness of our E-ARM for ameliorating the long-range coherence of generations, we undertake an experiment to assess the model's performance under different test sets with varying sentence lengths. We divided the test set of IWSLT14 (German $\rightarrow$ English, Italian $\rightarrow$ English, Spanish $\rightarrow$ English) translation dataset into three subsets ([0, 25], [25, 50], and [50, $\infty$)) based on the target sentence lengths. Then, we incrementally applied scheduled sampling technique and our E-ARM above the base transformer network, and tested their performances on these three subsets. Generally, the subset of samples with longer target sentences ([50, $\infty$)) should have been more affected by the long-range incoherence problem (lower BLEU score). In practice, we uniformly applied label smoothing and beam searching (with 5 beams) strategy for all experiments in Table~\ref{tab:effect_of_length}.

\begin{table}[h]
\centering
\renewcommand{\arraystretch}{1.2}
\resizebox{0.75\linewidth}{!}{
\begin{tabular}{c|cc|ccc|c}
\toprule
\textbf{Translation} & \textbf{Scheduled} & \textbf{E-ARM} & \multicolumn{3}{c|}{\textbf{Target Sentence Length}} & \multirow{2}{*}{\textbf{All Test}} \\
\cline{4-6}
\textbf{Task}& \textbf{Sampling} & \textbf{Training} & \textbf{[0, 25)} & \textbf{[25, 49)} & \textbf{[50, \bm{$\infty$})} & \\
\midrule[1.2pt]
\multirow{3}{*}{\textbf{De}\bm{$\rightarrow$} \textbf{En}} & - & - & 37.72 \footnotesize$\pm$\footnotesize 0.04 & 33.24 \footnotesize$\pm$\footnotesize 0.06 & 30.86 \footnotesize$\pm$\footnotesize 0.07 & 34.61 \footnotesize$\pm$\footnotesize 0.08 \\		
 & \ding{52} & - & 38.20 \footnotesize$\pm$\footnotesize 0.07 & 33.76 \footnotesize$\pm$\footnotesize 0.03 & 31.08 \footnotesize$\pm$\footnotesize 0.06 & 35.10 \footnotesize$\pm$\footnotesize 0.04 \\
 & \ding{52} & \ding{52} & 38.37 \footnotesize$\pm$\footnotesize 0.06 & 33.92 \footnotesize$\pm$\footnotesize 0.09 & 31.43 \footnotesize$\pm$\footnotesize 0.04 & 35.36 \footnotesize$\pm$\footnotesize 0.05 \\
\hline		
\multirow{3}{*}{\textbf{It}\bm{$\rightarrow$} \textbf{En}} & - & - & 35.20 \footnotesize$\pm$\footnotesize 0.03 & 32.73 \footnotesize$\pm$\footnotesize 0.02	 & 26.86 \footnotesize$\pm$\footnotesize 0.05 & 32.29 \footnotesize$\pm$\footnotesize 0.03 \\
 & \ding{52} & - & 35.52 \footnotesize$\pm$\footnotesize 0.09  & 33.25 \footnotesize$\pm$\footnotesize 0.08  & 26.95 \footnotesize$\pm$\footnotesize 0.14  & 32.64 \footnotesize$\pm$\footnotesize 0.12 \\
 & \ding{52} & \ding{52} & 35.56 \footnotesize$\pm$\footnotesize 0.10 & 33.33 \footnotesize$\pm$\footnotesize 0.13 & 27.21 \footnotesize$\pm$\footnotesize 0.07 & 32.82 \footnotesize$\pm$\footnotesize 0.11 \\
\hline
\multirow{3}{*}{\textbf{Es}\bm{$\rightarrow$} \textbf{En}} & - & - & 43.37 \footnotesize$\pm$\footnotesize 0.05 & 39.67 \footnotesize$\pm$\footnotesize 0.08 & 37.14 \footnotesize$\pm$\footnotesize 0.06 & 40.64 \footnotesize$\pm$\footnotesize 0.07 \\
 & \ding{52} & - & 43.61 \footnotesize$\pm$\footnotesize 0.09 & 40.00 \footnotesize$\pm$\footnotesize 0.04 & 37.38 \footnotesize$\pm$\footnotesize 0.06 & 40.91 \footnotesize$\pm$\footnotesize 0.06 \\		
 & \ding{52} & \ding{52} & 43.84 \footnotesize$\pm$\footnotesize 0.10 & 40.35 \footnotesize$\pm$\footnotesize 0.05 & 38.07 \footnotesize$\pm$\footnotesize 0.04 & 41.58 \footnotesize$\pm$\footnotesize 0.07 \\
\bottomrule
\end{tabular}
}
\caption{\footnotesize{Performance comparison on the IWSLT14 test set with respect to the different lengths of sentences on three translation tasks (German to English, Italian to English, and Spanish to English). Performance is evaluated by BLEU score.}}
\label{tab:effect_of_length}
\end{table}

Specifically, Table~\ref{tab:effect_of_length} shows that the base translation model improved performance for all three test sets with varying target sentence lengths after using the scheduled sampling technique, especially for the two sets [0, 25) and [25, 50) which had relatively short target sentence lengths (\textit{e.g.} On German to English task, 38.20 - 37.72 = +0.48 points and 33.76 - 33.24 = + 0.52 points for [0, 25) and [25, 50) test sets respectively). We consider that this performance boost was achieved through alleviating the exposure bias problem, since scheduled sampling approaches~\citep{SSRanzatoCAZ15, BridginfGapNMTijcai, SSTransformerMihaylovaM19} 
have been verified in mitigating the exposure bias problem. 
Besides, after applying our E-ARM together with the scheduled sampling technique, the base model can further obtain additional performance gain. Concretely, the improvement on the longer sentence is more evident, since model can obtain large improvements on the [50, $\infty$) (\textit{e.g.} On German to English task, 31.43 - 31.08 = +0.35 points for [50, $\infty$) test sets) than short sets [0, 25] and [25, 50] (\textit{e.g.} On German to English task, 38.37 - 38.20 = +0.17 points and 33.92 - 33.76 = + 0.16 points for [0, 25) and [25, 50) test sets respectively).  This phenomenon indicates that our E-ARM can resolve the incoherence problem to some extent.


\subsection{Effect on Exposure Bias}
\begin{table}[h]
\centering
\renewcommand{\arraystretch}{1.2}
\resizebox{0.8\linewidth}{!}{
\begin{tabular}{c|cccccc}
\toprule
\textbf{Trans. Pairs}& \textbf{DE$\rightarrow$ EN} & \textbf{EN$\rightarrow$ DE} & \textbf{EN$\rightarrow$ IT} & \textbf{IT$\rightarrow$ EN} & \textbf{ES$\rightarrow$ EN} & \textbf{EN$\rightarrow$ ES} \\
\midrule[1.2pt]
\bm{$\mathcal{N}$} & 14203 & 14554 & 14976 & 13952 & 16021 & 15359 \\
\textbf{Total} & 22148 & 23057 & 23654 & 23744 & 23860 & 22775 \\
\textbf{Ratio} & 64.12\% & 63.12\% & 63.31\% & 59.76\% & 68.33\% & 67.43\% \\
\bottomrule
\end{tabular}
}
\caption{\footnotesize{The effect of E-ARM on the exposure bias problem. Each test set of translation tasks contains 1K sentences selected randomly. $\mathcal{N}$ denote the ground truth words whose probabilities in the predicted distributions produced by E-ARM are greater than those produced by the baseline. }}
\label{tab:effect_exposure_bias}
\end{table}

We follow the analytic experiments in the work~\citep{BridginfGapNMTijcai} to show that our E-ARM is capable of alleviating the exposure bias problem. To be concrete, we randomly select 1K pairs from the training data for each translation pair and use the trained autoregressive model which applied E-ARM (Label Smoothing with smoothing factor 0.1 is applied during training while scheduled sampling is not used) to decode the source sentences, and then count the ground truth words whose probabilities in the predicted distributions produced by our E-ARM are greater than those produced by the baseline and denote the number as $\mathcal{N}$. The ratio of $\mathcal{N}$ to the total number of words tested is calculated. The detailed results are shown in Table \ref{tab:effect_exposure_bias}. We find that the results on all different tasks are greater than 50\%, which demonstrate the ability of our E-ARM in solving exposure bias problem.
\begin{table}[htp]
\centering
\renewcommand{\arraystretch}{1.2}
\resizebox{0.8\linewidth}{!}{
\begin{tabular}{c|c|cccccc}
\toprule
\multicolumn{2}{c|}{\textbf{Trans. Pairs}}& \textbf{DE$\rightarrow$ EN} & \textbf{EN$\rightarrow$ DE} & \textbf{EN$\rightarrow$ IT} & \textbf{IT$\rightarrow$ EN} & \textbf{ES$\rightarrow$ EN} & \textbf{EN$\rightarrow$ ES} \\
\midrule[1.2pt]
\multirow{3}{*}{\bm{$k$}} & \bm{$0$} & 34.93 & 28.91 & 30.04 & 32.56 & 41.01 & 37.73 \\
& \bm{$5$} & \textbf{34.97} & 28.92 & \textbf{30.08} & \textbf{32.60} & \textbf{41.07} & 37.71 \\
& \bm{$10$} & 34.95 & \textbf{28.95} & 30.07 & 32.59 & 41.03 & \textbf{37.75} \\
\bottomrule
\end{tabular}
}
\caption{\footnotesize{The effect of Top-K correction in the inference stage. We tested BLEU scores of using different $k$ on different translation pairs of IWSLT14 dataset. }}
\label{tab:effect_TopK}
\end{table}

\subsection{Analysis to Model's Convergence}
\begin{figure}[htp]
    \centering 
    \includegraphics[width=0.9\linewidth]{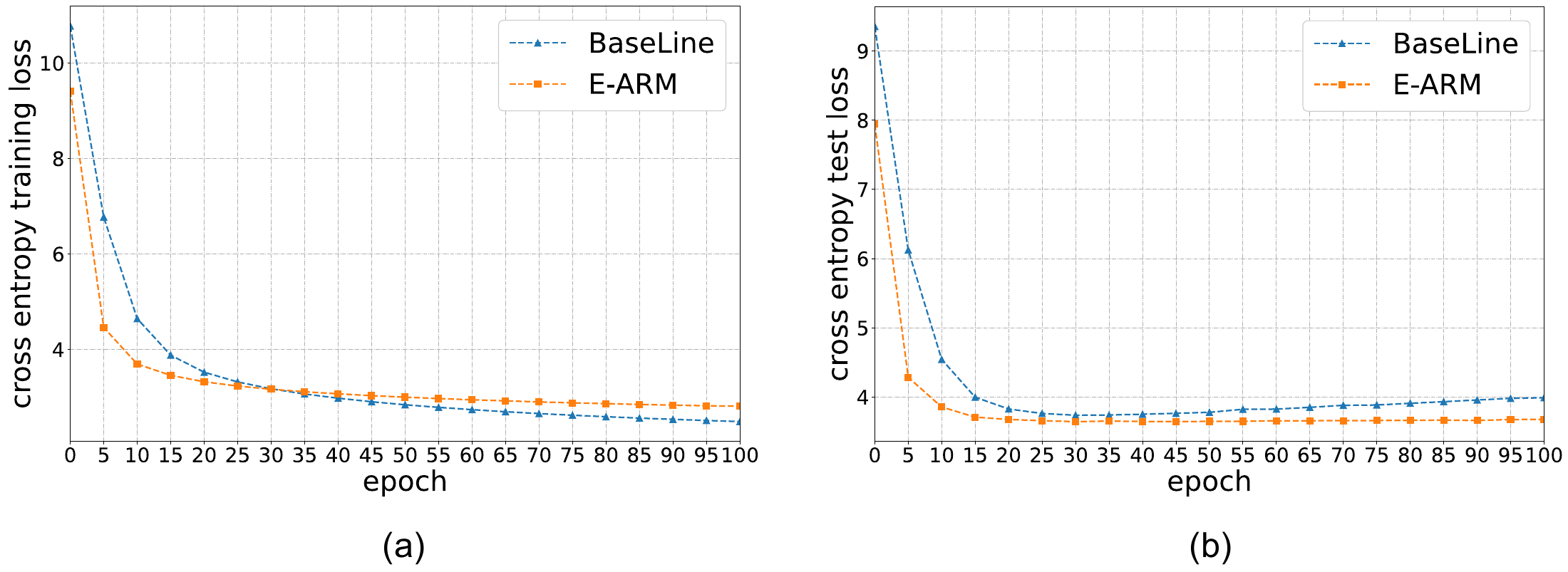}
    \caption{\footnotesize{(a) Cross entropy loss curves on IWSLT14 Spanish to English translation task on training set. The blue and orange colors represent base model and E-ARM respectively; (b) Cross entropy loss curves on IWSLT14 Spanish $\rightarrow$ English translation task on test set.}}
    \label{fig:instability_curves} 
\end{figure}

In this section, we will investigate the convergence of our E-ARM. To begin, we first train a base Transformer model (``Tr-Base'' architecture shown in Table \ref{tab:hyper-parameter}) on the IWSLT14 Spanish to English training set for baseline and E-ARM model respectively, and then record the training loss and test loss (in cross entropy) at the end of each epoch. The loss curves are plotted in the Figure~\ref{fig:instability_curves}. From Figure~\ref{fig:instability_curves}, we can see that (1) at the start of the training, our E-ARM converges slightly faster than the baseline. (2) As the training process progresses, the cross entropy of the baseline on the training set will gradually decrease, with a faster rate than E-ARM. On the other hand, the test loss curve of the baseline will fall at initially and then slowly rise after 50 epochs while E-ARM always remains stable convergence. This phenomenon also shows that our E-ARM model can effectively prevents over-fitting so that obtaining better generalization.

\begin{table}[htp]
\centering
\renewcommand{\arraystretch}{1.3}
\resizebox{0.9\linewidth}{!}{
\begin{tabular}{c|cc|ccccc}
\toprule
\multirow{2}{*}{\textbf{Trans. Pairs}} & \textbf{Scheduled} & \textbf{E-ARM} & \multicolumn{5}{c}{\textbf{Metrics}} \\
\cline{4-8}
& \textbf{Sampling} & \textbf{Training} & \textbf{ROUGE-1} \bm$\uparrow$ & \textbf{ROUGE-2}\bm$\uparrow$ & \textbf{ROUGE-L}\bm$\uparrow$ & \textbf{METEOR}\bm$\uparrow$ & \textbf{BLEU}\bm$\uparrow$ \\
\midrule[1.2pt]
\multirow{3}{*}{\textbf{De} \bm{$\rightarrow$} \textbf{En}} & - & - & 66.51 & 43.69 & 63.69 & 64.35 & 34.61  \\		
 & \ding{52} & - & 66.83 & 44.08 & 64.02 & 64.61 & 35.10  \\
 & \ding{52} & \ding{52} & \textbf{67.46} & \textbf{44.77} & \textbf{64.78} & \textbf{65.13} & \textbf{35.36} \\
\hline
\multirow{3}{*}{\textbf{It} \bm{$\rightarrow$} \textbf{En}} & - & -  & 64.50 &	40.65 & 61.69 & 62.18 & 32.29 \\		
 & \ding{52} & - &  64.73 & 40.97 & 61.94 & 62.51 & 32.64 \\
 & \ding{52} & \ding{52} & \textbf{65.27} & \textbf{41.51} & \textbf{62.49} & \textbf{62.80} & \textbf{32.82} \\
\hline
\multirow{3}{*}{\textbf{Es} \bm{$\rightarrow$} \textbf{En}}  & - & - & 71.10 & 49.47 & 68.78 & 68.94 & 40.64 \\
 & \ding{52} & - & 71.36 & 49.53 & 68.96 & 69.28 & 40.91 \\
 & \ding{52} & \ding{52} & \textbf{71.91} & \textbf{50.17} & \textbf{69.65} & \textbf{69.63} & \textbf{41.58} \\
\bottomrule
\end{tabular}
}
\caption{\footnotesize{Comparison of ROUGE-1, ROUGE-2, ROUGE-L, METEOR, and BLEU scores between our approach E-ARM and the base ARGM trained just with cross-entropy loss on three translation pairs of IWSLT14 datasets. The value is expressed in percentage. We use ``Tr-Base'' as the network architecture.}}
\label{tab:other_metrics}
\end{table}
\subsection{Analysis to Top-K Energy Re-sampling}
Top-K energy re-sampling in the inference stage is introduced by \citet{ResidualBakhtinDGORS21}, which collects many candidate sequences generated autoregressively in the inference stage and then re-samples from them depending on their energy scores estimated by the network. To measure the effectiveness of the Top-K energy re-sampling towards our method, we conduct ablation study on neural machine translation task by selecting different K = \{0, 5, 10\}. The results are shown in Table~\ref{tab:effect_TopK} and performances are evaluated by using BLEU score. From Table~\ref{tab:effect_TopK}, we observe that the benefits brought by Top-K sampling is minor (K=\{5, 10\}), when compared with model without Top-K sampling (K=0). This together with the results shown in Table \ref{tab:language_modeling} shows that our E-ARm can considerably benefit the base autoregressive model even without the energy resampling technique.


\subsection{Evaluation with Other Metrics}
To further evaluate the effectiveness of the our proposed E-ARM, we also evaluate our method by using other metrics, such as ROUGE~\citet{lin2004rouge} and METEOR~\cite{banerjee2005meteor} for neural machine translation. The results are shown in Table~\ref{tab:other_metrics}. In Table~\ref{tab:other_metrics}, the improvements of E-ARM in different metrics is consistent with the conclusion of Table~\ref{tab:iwslt14-6PAIRS}, which further prove the effectiveness of our E-ARM model.

\subsection{Cases Studies}
To better understand the advantages of our method in correcting error tokens, we also prepare some translation cases in IWSLT14 German $\rightarrow$ English, as shown in Table~\ref{tab:case_study}.
\begin{table}[h]
\tiny
\begin{tabularx}{\textwidth}{X|X}
\toprule
\textbf{Source Sentence(German)} &\textbf{Predicted Target Sentence(English)} \\
\midrule[1.2pt]
wenn ich ihnen 600 zeitschriften zeige und sie in 10 kategorien aufteile oder ich ihnen 400 zeitschriften zeige, und diese in 20 kategorien aufteile, dann glauben sie, dass ich ihnen mehr auswahl und eine bessere auswahlerfahrung gegeben habe, als ich ihnen die 400 gegeben hätte gegenüber dem, wenn ich ihnen die 600 gegeben hätte. & \textbf{GroundTruth}: if i show you 600 magazines and i divide them up into 10 categories, versus i show you 400 magazines and divide them up into 20 categories, you believe that i have given you more choice and a better choosing experience if i gave you the 400 than if i gave you the 600.  \\
& \textbf{Baseline}: if i show you 600 magazines and i split them in 10 categories, or i'm showing them 400 magazines, and i'm going to split them up into 20 categories, you think i've given them more choices and better choice than i would have given them the 400 over the time that i gave them the 600. \\
& \textbf{Baseline + S.S.}:  if i show you 600 magazines and i give you 400 magazines in 10 categories, and i give you 400 magazines, and i can split them up in 20 categories, then you think i've given you more choice and a better selection than i would have given you the 400 of which if i gave you the 600. \\
& \textbf{Ours}: if i show you 600 magazines and i divide them into 10 categories, or i show you 400 magazines, and i divide them into 20 categories, you think i've given you more choices and better selection experience than i gave you the 400 of whom if i gave you the 600.\\
\hline
und ich weiß definitiv, dass es für mich – in meiner situation – sehr gefährlich wäre, anzufangen, diesen dunklen pfad der vermutung sozusagen herunterzusickern – besonders in dem umstand, in dem ich mich in meiner karriere gerade befinde. & \textbf{GroundTruth}: and i definitely know that, in my case -- in my situation -- it would be very dangerous for me to start sort of leaking down that dark path of assumption, particularly given the circumstance that i'm in right now in my career. \\
 & \textbf{Baseline}: and i know definitely, for me, it would be very dangerous to begin to do this dark path of suspect -- especially in the circumstance that i'm in my career right now. \\
 & \textbf{Baseline + S.S.}: and i know definitely it would be -- in my situation -- very dangerous to start, to kind of settle down this dark path of presumption -- especially in the circumstance in which i'm in my career right now. \\
 & \textbf{Ours}: and i definitely know that it's for me -- in my situation -- very dangerous to start to sickle down this dark path of suspection, in particular, in the circumstance of where i'm in my career right now. \\
\hline
wir haben das licht ausgeschaltet, legten es in ein vakuum und saugten die ganze luft aus und kühlten es bis fast zum jetzt, ganz alleine im aufzug, war das stück metall frei, sich zu verhalten wie immer es wollte. & \textbf{GroundTruth}: we turned off the lights, and then we put it in a vacuum and sucked out all the air, and then we cooled it down now, all alone in the elevator, the little chunk of metal is free to act however it wanted.\\
& \textbf{Baseline}: we turned the light off, put it in a vacuum and sucked it out all the air and cooled it up until almost now, all the way alone, the piece of metal was open to behave as it was.\\
& \textbf{Baseline + S.S.}: we turned the lights off, we put it into a vacuum, and we sucked all the air, and we cooled it all the way up to now, all over the place, the piece of metal was free to behave whatever it wanted. \\
 & \textbf{Ours}: we turned off the lights, we put it into a vacuum and we sucked all the air out, and we cooled it up until almost now, all alone in the elevator, the piece of metal was free to behave whatever it wanted. \\
\hline
und im grunde können sie das betrachten, wissen sie, als eine tyrannei des erinnernden selbst, und sie können sich das erinnernde selbst denken als eins, das sozusagen das erlebende selbst schleppt durch erfahrungen, die das erlebende selbst nicht braucht. & \textbf{GroundTruth}: and basically you can look at this, you know, as a tyranny of the remembering self, and you can think of the remembering self sort of dragging the experiencing self through experiences that the experiencing self doesn't need. \\
& \textbf{Baseline}: and basically, you can think of this, you know, as a tyranny of self, and you can think of the memorable self as one that kind of weaves the living self through experiences that don't need the life itself. \\
& \textbf{Baseline + S.S.}: and basically, you can look at this, you know, as a tyrannei of memorial self, and you can think of the memorial self as one that kind of sucks the living self through experiences that don't need the living self. \\
& \textbf{Ours}: and basically, you can look at that, you know, as a tyranny of the remembering self, and you can think of the memory itself as one, which is sort of dragging the living self through experiences that the living self doesn't need. \\
\hline
wir sind an der schwelle zu erstaunlichen, erstaunlichen ereignissen auf vielen gebieten. und doch denke ich wirklich, dass wir hunderte, 300 jahre vor die aufklärung zurück gehen müssten, um eine zeit zu finden, in der wir fortschritt bekämpft haben, in der wir über diese dinge heftiger getritten haben, an mehr fronten als jetzt.& \textbf{GroundTruth}: we're on the verge of amazing, amazing events in many fields, and yet i actually think we'd have to go back hundreds, 300 years, before the enlightenment, to find a time when we battled progress, when we fought about these things more vigorously, on more fronts, than we do now. \\
& \textbf{Baseline}: we are at the threshold of amazing, amazing events in many areas, and yet i really think that we have to go back hundreds and 300 years before the enlightenment to find a time when we have fought progress in which we have driven more of these things than now. \\
& \textbf{Baseline + S.S.}: we're at the threshold of amazing, amazing events in many areas. and yet, i really think that we have to go back hundreds and hundreds of years before the enlightenment to find a time when we have struggled with progress in which we have driven on these things more powerful, more fronts than now. \\
& \textbf{Ours}: we're at the threshold to amazing, amazing events in many areas, and yet i really think that we have to go back hundreds and 300 years before the enlightenment to find a time when we fought progress, where we've been fighting about these things to more fronts than we have now. \\
\bottomrule
\end{tabularx}
\tabcaption{\footnotesize{Translation cases on IWSLT14 De$\rightarrow$En test set, generated by the baseline method, baseline with scheduled sampling and our
E-ARM. The italic font means the mismatch translation}}
\label{tab:case_study}
\end{table}



\end{document}